\documentclass[11pt]{article}

\usepackage[final]{acl}

\usepackage{times}
\usepackage{latexsym}

\usepackage[T1]{fontenc}

\usepackage[utf8]{inputenc}

\usepackage{microtype}

\usepackage{inconsolata}

\usepackage{graphicx}

\usepackage{amsmath}
\usepackage{booktabs}
\usepackage{eso-pic}
\title{BilliardPhys-Bench: Benchmarking Physical Reasoning and Visual Dynamics of Multimodal LLMs}

\author{
  Ben Wang\textsuperscript{*} \\
  Alibaba Group \\
  Beijing, China \\
  {\small\texttt{yuanjian.wb@alibaba-inc.com}}
  \And
  Xiaogang Li\textsuperscript{*} \\
  Alibaba Group \\
  Beijing, China \\
  {\small\texttt{lixiaogang.lxg@alibaba-inc.com}}
  \And
  Ruochen Gao \\
  Alibaba Group \\
  Beijing, China \\
  {\small\texttt{gaoruochen.grc@alibaba-inc.com}}
  \AND
  Peiyao Xiao \\
  Alibaba Group \\
  Beijing, China \\
  {\small\texttt{xiaopeiyao.xpy@alibaba-inc.com}}
  \And
  Chengliang Xu \\
  Alibaba Group \\
  Beijing, China \\
  {\small\texttt{xiaodu.xcl@alibaba-inc.com}}
  \And
  Zeyu Wang \\
  Alibaba Group \\
  Beijing, China \\
  {\small\texttt{chenfan.wzy@alibaba-inc.com}}
  \AND
  Zichao Chen \\
  Alibaba Group \\
  Beijing, China \\
  {\small\texttt{chenzichao.czc@alibaba-inc.com}}
  \And
  Bing Zhao\textsuperscript{\textdagger} \\
  Alibaba Group \\
  Beijing, China \\
  {\small\texttt{xiongdao@alibaba-inc.com}}
  \And
  Hu Wei\textsuperscript{\textdagger} \\
  Alibaba Group \\
  Beijing, China \\
  {\small\texttt{kongwang@alibaba-inc.com}}
}

\begin{document}

\AddToShipoutPictureBG*{%
  \AtPageUpperLeft{%
    \put(60,-45){\includegraphics[height=0.7cm]{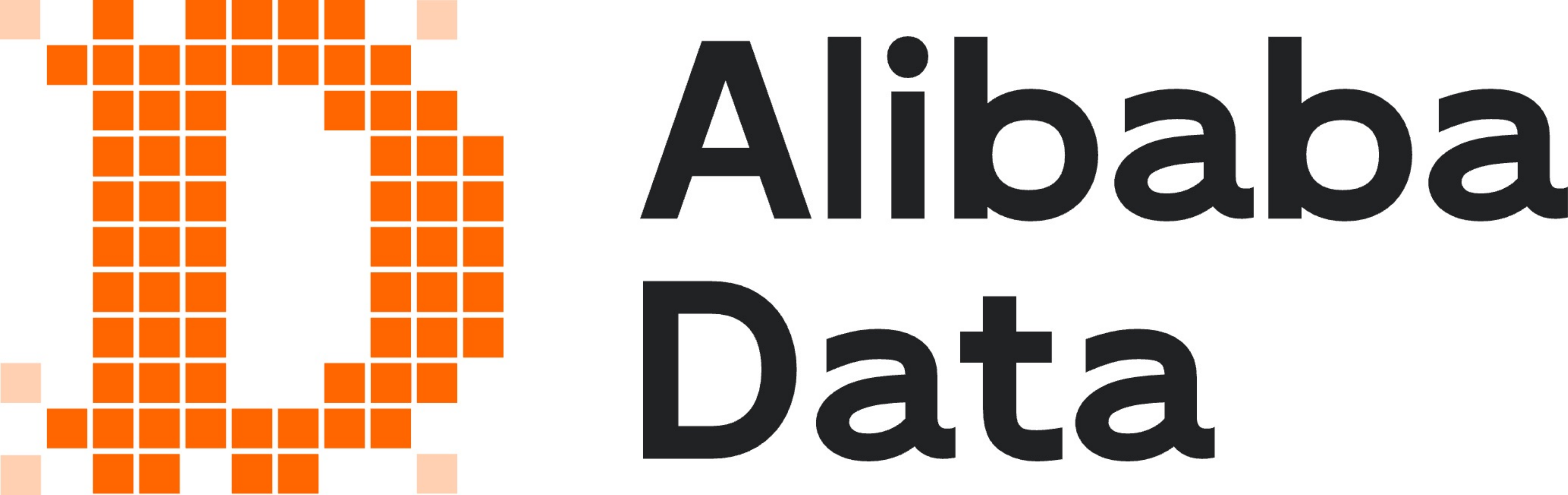}}%
  }%
  \AtPageUpperLeft{%
    \put(400,-45){\includegraphics[height=0.8cm]{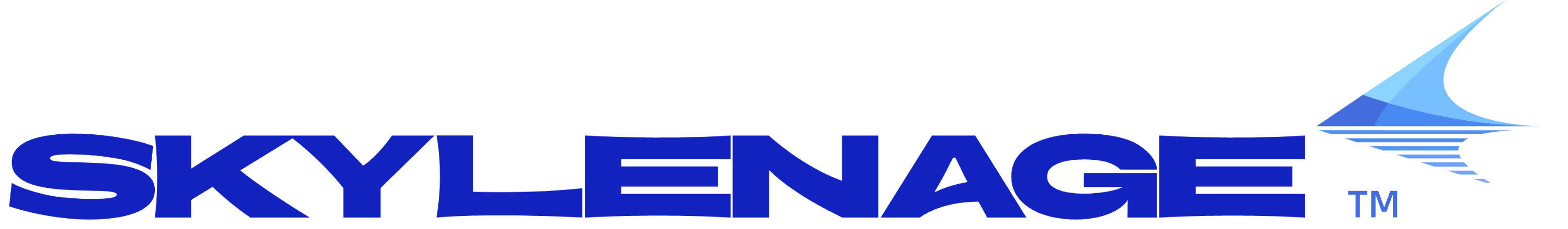}}%
  }%
  \AtPageUpperLeft{%
    \put(49,-56){\rule{500pt}{0.4pt}}%
  }%
}

\maketitle

\begin{abstract}
Current multimodal models handle static image recognition well, but intuitive physical reasoning remains a weakness. Predicting how objects will move and interact from a single image is still difficult for these systems.
We present BilliardPhys-Bench, a benchmark for physical reasoning in synthetic billiards environments. Its procedural engine generates randomized scenarios with friction and elastic collisions. The benchmark tests three abilities: (1) predicting ball-to-ball collisions, (2) reasoning about wall bounces, and (3) estimating final ball positions after motion stops.
We evaluate recent MLLMs from the GPT, Claude, Gemini, and Qwen families. Performance drops as simulation time increases and scene geometry grows more complex. We also observe a consistent failure mode we call "stasis bias": when the correct physical outcome is harder to infer, models tend to predict no interaction. These findings show where current MLLMs break down on visual dynamics and point toward the need for better physical inductive biases in multimodal architectures.
\end{abstract}

\section{Introduction}

Reasoning about the physical world is a prerequisite for general-purpose AI \cite{Agarwal2025,wu2024,Upadhyay2026,zhao2026,zhou2026}. Humans do this easily: from a single image of a billiard table, most people can anticipate likely trajectories, collisions, and outcomes. Current Multimodal Large Language Models (MLLMs), however, still struggle to infer accurate physical evolution from a 2D image \cite{Yao2024,Ye2024,2026arXiv260110384Z,zhang2026-1}.

Billiards is well-suited for evaluation because it requires geometric estimation of collision angles, multi-body tracking, and causal reasoning over dynamics \cite{zhao2015,kief2024pooltool,2025arXiv250607860A}. Unlike standard visual question answering, which mainly tests semantic recognition, billiard prediction is sensitive to initial conditions: small errors in perceived position or velocity compound into substantially different outcomes \cite{wang2025physunibench,xu2025,2025arXiv251216969X,2026arXiv260118744Y,2026arXiv260116007M}. This sensitivity makes billiards a test of predictive physical reasoning rather than descriptive scene understanding \cite{yuan2025being,2025arXiv251211047J,2025arXiv251003081L}.

Existing benchmarks for physical reasoning fall into three categories. Static QA benchmarks test knowledge of physical laws through text or image queries \cite{wang2025physunibench,shen2025phyxdoesmodelwits,xiang2025seephys,Chung:2025nsd,2025arXiv250926574Z,2025arXiv251024591Y,2026arXiv260114235L}---these measure declarative knowledge but do not require reasoning about evolving environments. Symbolic or code-based benchmarks present physics through abstract interfaces, reducing the need for perceptual grounding \cite{bakhtin2019phyrenewbenchmarkphysical,matthews2025kinetixinvestigatingtraininggeneral,cherian2024llmphycomplexphysicalreasoning,2025arXiv251123465L,2025arXiv250704766Z}. Visual interactive benchmarks evaluate VLMs as agents that plan and adapt through trial and error \cite{2025arXiv251123465L,2025arXiv250704766Z}. Single-frame forward simulation---predicting future states from a static initial condition without iterative interaction---remains less studied.

This gap reflects a separation between perception and reasoning. Models may describe scenes or state physical rules correctly yet fail to produce accurate predictions \cite{xu2025deepphybenchmarkingagenticvlms,2025arXiv251223292L,2025arXiv251219526P,2025arXiv251117649L}. In billiards, this appears as stasis bias and failures on sequential collisions. Prior work on ball trajectory prediction has mainly used frame-by-frame tracking in video \cite{yu2025fast,kienzle2025upliftingtabletennisrobust,gomezgonzalez2020realtimetrajectoryprediction,2024ITPro..26c..65C} rather than prediction from a static state.

We introduce BilliardPhys-Bench with three contributions:

(1) A procedurally generated, multi-tier benchmark. Our engine creates randomized billiard scenarios, each providing a high-resolution initial-state image paired with physics-engine ground truth. The benchmark decomposes the problem into three levels: (a) discrete collision prediction (will the cue ball collide, and with which object?), (b) continuous final-state estimation (precise coordinates of all balls after energy dissipates), and (c) complex interaction chains (reasoning through secondary and tertiary collisions).

(2) A diagnostic framework for physical reasoning failures. We evaluate leading proprietary and open-source MLLMs and analyze failure modes---stasis bias, misunderstanding of momentum transfer, sensitivity to visual details---providing specific insight into why current architectures fail at visual dynamical reasoning.

(3) Directions for stronger physical inductive biases. Our results show the limits of next-token prediction for physical simulation and outline pathways through hybrid neural-physical approaches \cite{achterhold2023blackboxvsgrayboxcase}, factor graphs, or differentiable physics engines.

By shifting evaluation from interactive action to one-shot prediction, our benchmark tests whether MLLMs can maintain an internal model of physics.

\section{Data Generation}
\begin{figure}[htbp]
    \centering
    \includegraphics[width=0.5\textwidth]{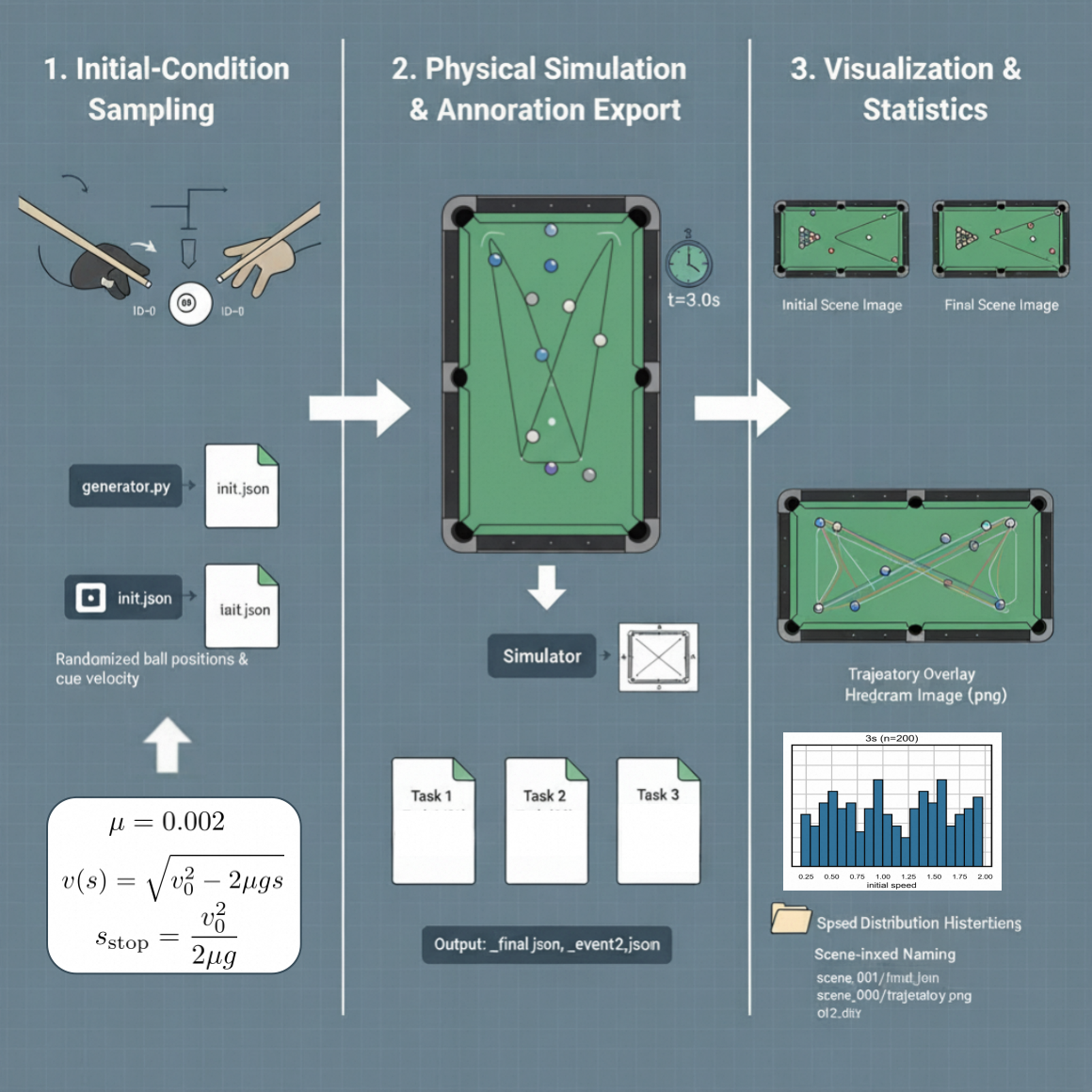}
    \caption{BilliardPhys-Bench data generation pipeline. The pipeline has three stages: (1) initial-condition sampling, which randomizes ball positions and cue velocities; (2) physical simulation, in which a high-fidelity engine computes elastic collisions and friction-based deceleration while exporting structured annotations; and (3) visualization and statistics, which generate the rendered images and dataset distribution reports.}
    \label{fig:dataflow}
\end{figure}
\subsection{Physical modeling of constant friction}

To describe the motion of a billiard ball under constant friction, we consider a ball of mass $m$ moving on a horizontal surface with a coefficient of friction $\mu$. The kinetic friction force $f$ acting against the direction of motion is:
\begin{equation}
    f = \mu N = \mu mg
\end{equation}
where $N$ is the normal force and $g$ is gravitational acceleration. By Newton's second law ($F = ma$), the constant deceleration $a$ is:
\begin{equation}
    a = \frac{-f}{m} = \frac{-\mu mg}{m} = -\mu g
\end{equation}
For uniformly decelerated linear motion, the relationship between final velocity $v$, initial velocity $v_0$, acceleration $a$, and displacement $s$ is:
\begin{equation}
    v^2 - v_0^2 = 2as
\end{equation}
Substituting $a = -\mu g$ gives velocity as a function of displacement:
\begin{equation}
    v^2 = v_0^2 - 2\mu gs, \qquad v(s) = \sqrt{v_0^2 - 2\mu gs}
    \label{eq:standard_friction}
\end{equation}
Under this constant friction model, ball velocity decreases non-linearly with distance traveled (Eq.~\ref{eq:standard_friction}).

\subsection{Dataset generation}
The dataset was synthesized using a custom generator. Each example is created through three stages: initial-condition sampling, physical simulation and annotation export, and visualization (see Figure~\ref{fig:dataflow}).

First, an initial scene is sampled and saved as \texttt{init.json}: cue ball (ID=0) velocity (direction and magnitude) and positions of all other balls are randomly sampled according to configured ranges and seed. Friction is modeled as speed decrease proportional to path length (Eq.~\ref{eq:standard_friction}), so a ball stops after traveling:
\begin{equation}
    s_{\text{stop}} = \frac{v_0^2}{2 \mu g}.
\end{equation}
Here $g = 9.8 \, \mathrm{m/s^2}$ and $\mu = 0.002$. We chose this friction coefficient so that collisions occur in roughly 50\% of generated scenarios given our initial velocity intervals. All collisions (ball-ball and ball-wall) are perfectly elastic. If a ball is pocketed during simulation, its final position is recorded as \texttt{null}.
\begin{figure}[htbp]
    \centering
    \includegraphics[width=0.45\textwidth]{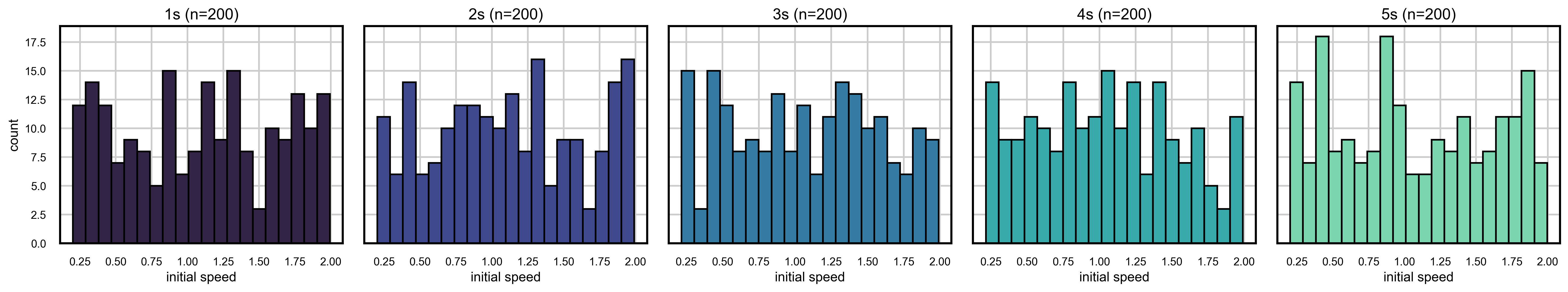} 
    
    \includegraphics[width=0.45\textwidth,trim=1mm 10mm 1mm 1mm]{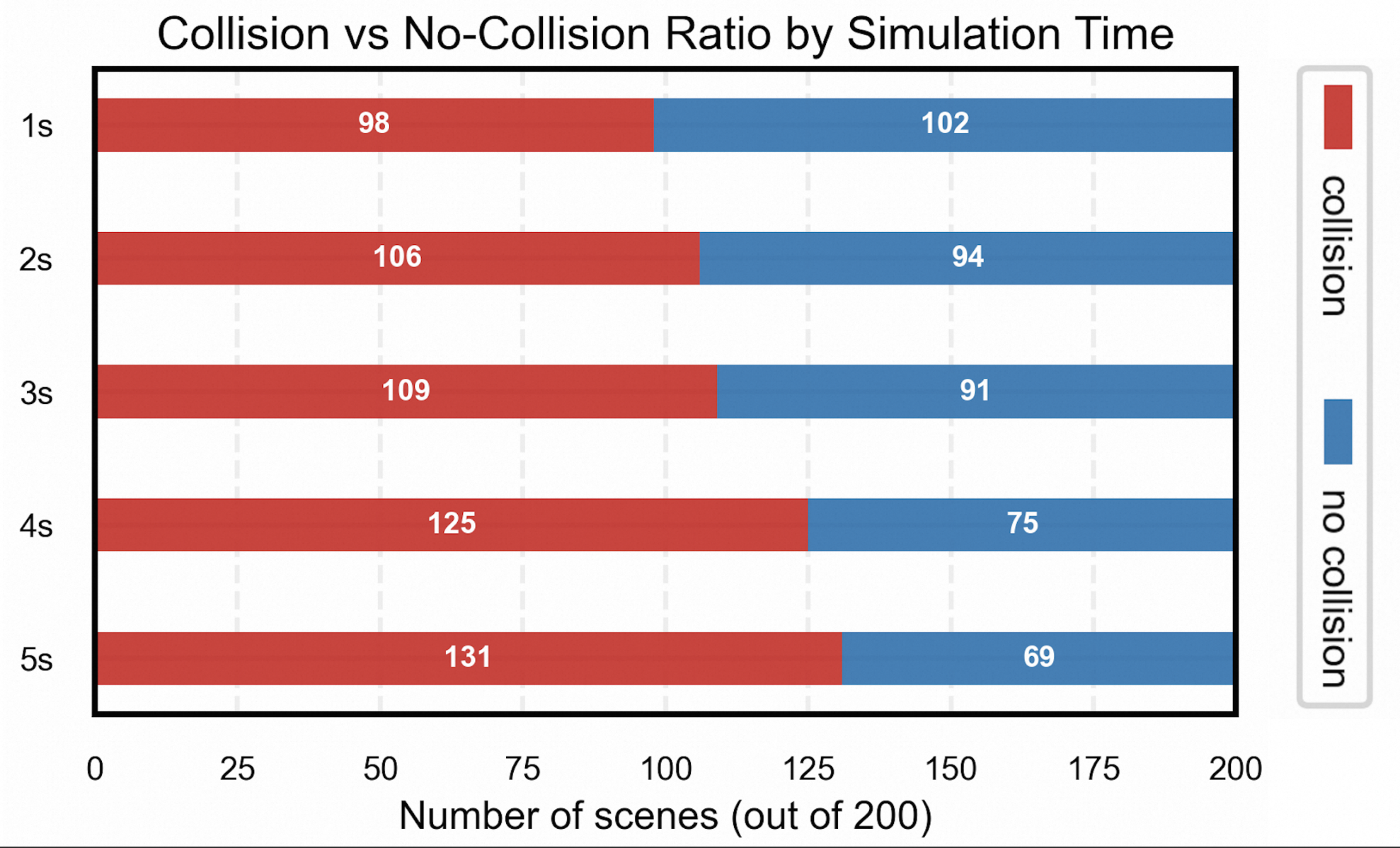}
    \caption{Distribution of the generated benchmark samples. The top histograms show sampled cue-ball initial speeds for simulation windows from 1 s to 5 s. The bottom bar chart shows the number of collision and no-collision cases among the 200 sampled scenes at each time window.}
    \label{fig:scenedis}
\end{figure}
Second, the simulator advances to the target time ($t \in \{1,2,3,4,5\}\,\text{s}$) and writes \texttt{\_final.json} with:

\begin{itemize}
    \item \textbf{Task 1 (Q1):} Per-ball collision labels saved as a list of objects in the form \texttt{\{"id": i, "answer": "T" | "F"\}}, with one entry per non-cue ball.
    \item \textbf{Task 2 (Q2):} Per-wall-collision answers stored as an answer list (e.g., \texttt{\{"wall": "TOP", "answer": "T" | "F"\}}) representing collision or no collision.
    \item \textbf{Task 3 (Predictions):} Final positions saved under the key \texttt{predictions} as a list of \texttt{\{"id": <int>, "pos": [x, y]\}}; pocketed balls use \texttt{pos = null}. Numeric coordinates are recorded to at most four decimal places.
\end{itemize}
\begin{figure*}[t]
    \centering
    \includegraphics[width=0.9\textwidth]{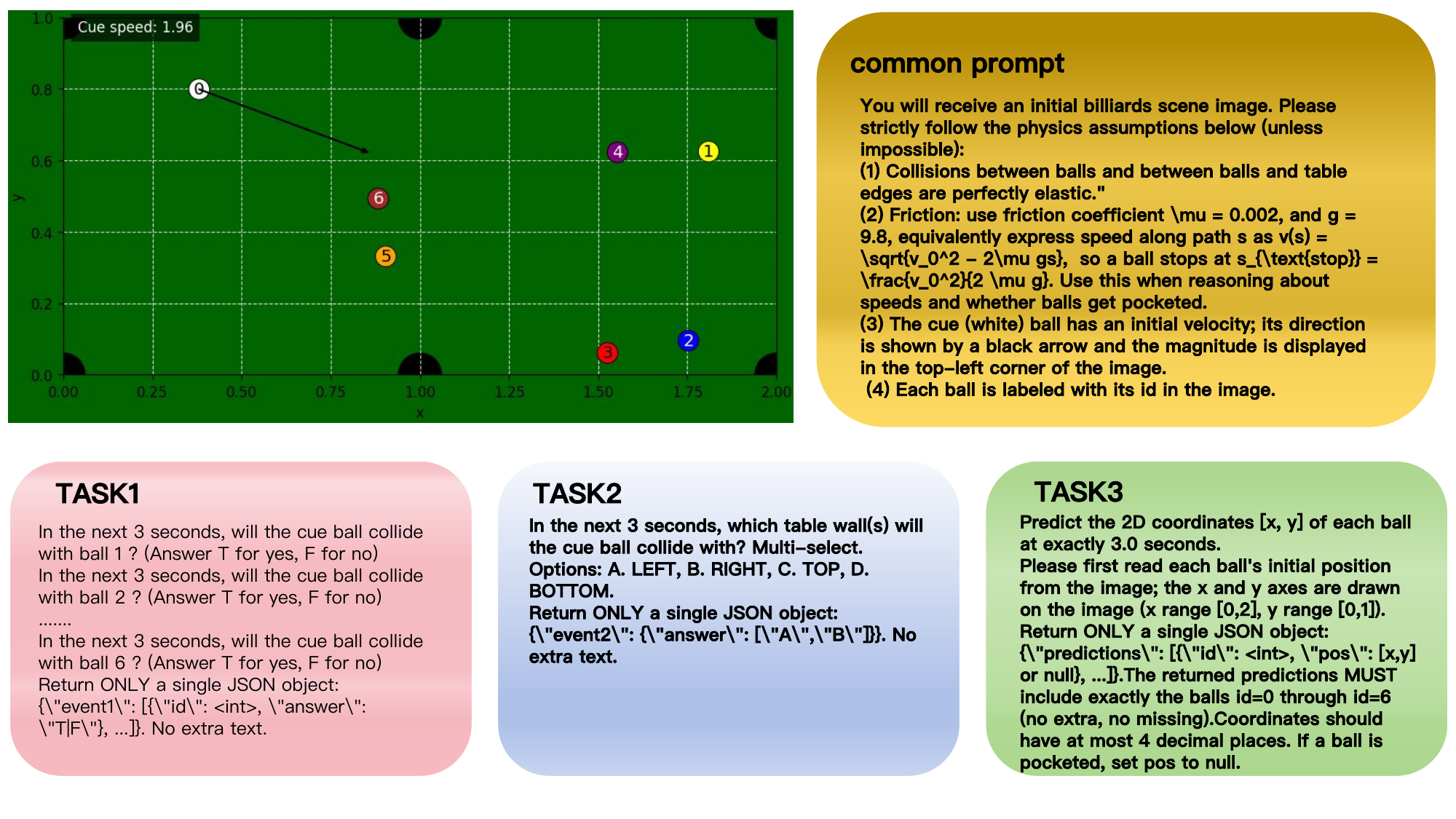}
    \caption{Illustration of the prompt engineering and task definitions. The evaluation pipeline provides models with a ``common prompt'' detailing physical constants (friction, elasticity) and the input image. The three diagnostic tasks, collision detection (Task 1), wall interaction (Task 2), and coordinate prediction (Task 3), require the model to return structured JSON outputs for automated scoring. This is an example for the 3s scenes.}
    \label{fig:question}
\end{figure*}
Third, the generator produces visual outputs: initial and final scene images, a trajectory overlay rendered with the same canvas size and ball radii as the frame renderer, and a histogram of sampled cue-ball initial speeds (Figure~\ref{fig:scenedis}). All files for a scene are stored under a scene-indexed naming convention for automated evaluation and reproducibility.

\subsection{Model I/O and evaluation pipeline}

The pipeline (Figure~\ref{fig:question}) constructs a structured prompt with an optional scene image, submits it to a chat-style LLM API, extracts JSON, and judges correctness against ground truth.

\begin{figure}[h]
    \centering
    \includegraphics[width=0.45\textwidth]{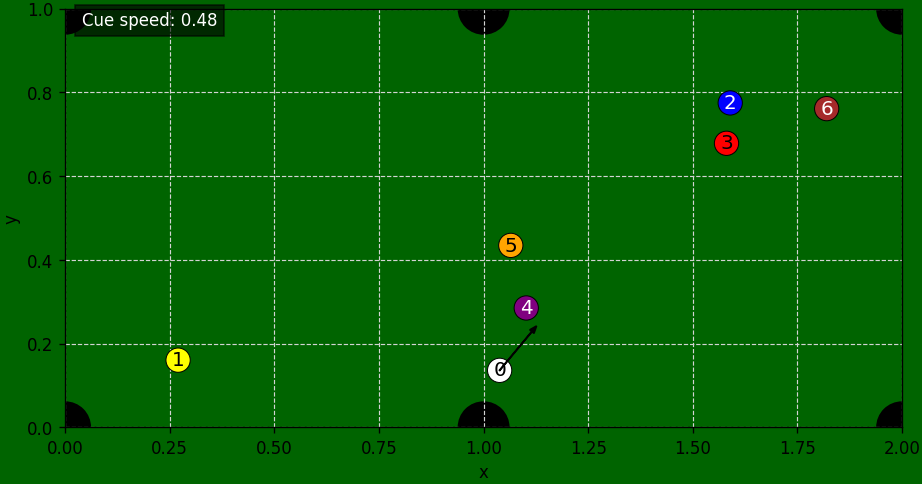} 

    \includegraphics[width=0.45\textwidth]{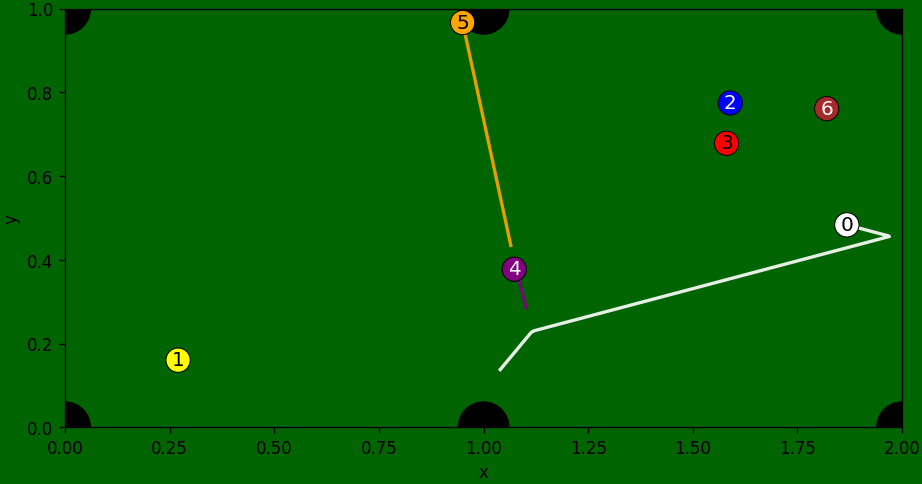} 

    \includegraphics[width=0.45\textwidth]{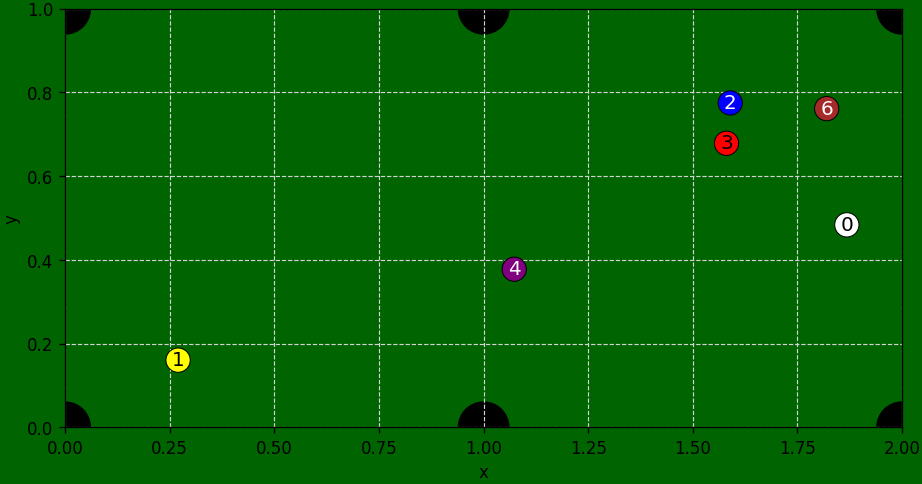} 
    \caption{Example from the billiard reasoning benchmark. (Top) The input image, with the initial ball positions and cue-ball velocity vector. (Middle) The ground-truth trajectory overlay, showing wall reflections and ball-to-ball collisions. (Bottom) The board state at the target prediction time.}
    \label{fig:billiard_scene}
\end{figure}

\paragraph{Inputs sent to the model}
\begin{itemize}
    \item \textbf{Image:} Scene image showing ball positions, IDs, and cue velocity arrow (Figure~\ref{fig:billiard_scene}), with axes $x \in [0,2], y \in [0,1]$.
    \item \textbf{Prompt:} States friction ($v(s) = \sqrt{v_0^2 - 2\mu gs}$, $\mu = 0.002$, $g = 9.8$), stopping distance ($s_{\text{stop}} = v_0^2 / (2\mu g)$), three task definitions, and JSON output format.
\end{itemize}

The three tasks:
\begin{enumerate}
    \item \textbf{Task 1 (Q1):} For each non-cue ball, predict whether the cue ball collides with it.
    \item \textbf{Task 2 (Q2):} For each wall, predict whether the cue ball collides with it.
    \item \textbf{Task 3 (Q3):} Predict the 2D position of every ball at the target time as \texttt{\{"id": <int>, "pos": [x, y]\}}; pocketed balls use \texttt{pos = null}.
\end{enumerate}

\paragraph{API call strategy}
The implementation uses separate model calls for Tasks 1, 2, and 3 to reduce output-format ambiguity and allow validation rules tailored to each task. All calls use the configured chat-completions endpoint with JSON-formatted messages. Chain-of-thought is enabled for Q1/Q2 but disabled for Q3, where free-form reasoning interferes with the required JSON output. Runtime parameters include \texttt{--max-tokens}, \texttt{--timeout}, and \texttt{--enable-cot}.

\paragraph{Validation and retry}
After each reply, the client extracts JSON by locating the first balanced \texttt{\{...\}} substring. Failed or truncated responses trigger retries with escalation strategies (fresh session or increased timeout).
\begin{itemize}
    \item \textbf{Task 1/2:} Parsed against expected schema; repeated if malformed or incomplete.
    \item \textbf{Task 3:} Must contain a \texttt{predictions} list covering IDs 0--6 with at least one non-null numeric coordinate and range checks. When \texttt{force\_numeric} is enabled, the prompt instructs numeric estimates rather than \texttt{null}.
\end{itemize}
Successful outputs are saved as three JSON files, one per task.

\paragraph{Correctness judgments}
A separate utility evaluates predictions against ground truth:
\begin{itemize}
    \item \textbf{Q1 (collisions):} T/F compared per ball against GT labels.
    \item \textbf{Q2 (walls):} Correct only if the predicted wall set exactly matches GT.
    \item \textbf{Q3 (positions):} Correct if Euclidean distance $\leq$ ball radius; both-null counts as correct.
\end{itemize}

\paragraph{Practical notes}
The prompt requires JSON-only outputs with explicit examples. We set temperature to 0 and impose no maximum token limit for all models. In base vs.\ ``Pro'' GPT comparisons, prompt format and client settings are fixed; observed token-usage differences reflect model-side behavior.

\section{Results}

Our evaluation on \texttt{BilliardPhys-Bench} reveals clear differences in how MLLMs handle physical reasoning. We analyze performance across three diagnostic tasks at temporal horizons from 1s to 5s.

\begin{figure*}[t]
    \centering
    \includegraphics[width=0.99\textwidth,trim=20mm 10mm 20mm 10mm]{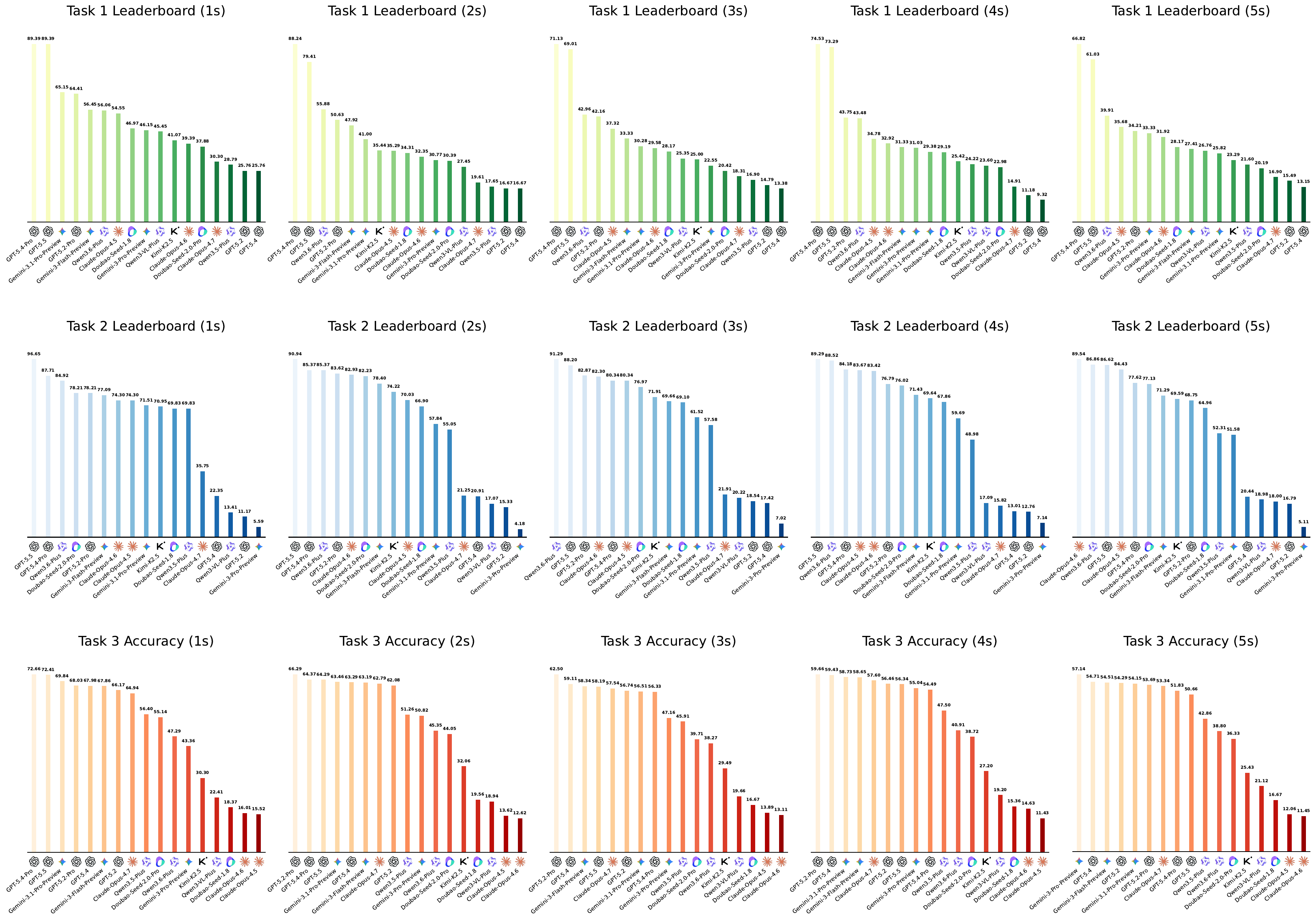} 
    \caption{Performance of leading MLLMs across the three tasks. The bar charts compare model accuracy as the simulation time increases from 1 s to 5 s. Accuracy decreases across all three tasks as the prediction window lengthens.}
    \label{fig:task_accuracy}
\end{figure*}

\subsection{Task 1: collision event prediction}
Task 1 separates models with event-level reasoning from those that only produce plausible final states (Figure~\ref{fig:task_accuracy}, Figure~\ref{fig:totalscore}).

\begin{itemize}
    \item \textbf{Top tier:} \textit{GPT-5.4-Pro} and \textit{GPT-5.5} form the first tier with mean accuracies of 78.02\% and 74.43\%, both remaining above 60\% even at 5 s. \textit{Qwen3.6-Plus} forms a competitive second tier, showing that strong event reasoning is no longer limited to one model family.
    
    \item \textbf{Structural weakness:} \textit{GPT-5.4} and \textit{Claude-Opus-4.7} achieve only 15.66\% and 20.01\% mean accuracy---weak from 1 s onward, indicating a basic deficit in discrete event identification rather than a long-horizon failure.
\end{itemize}

\subsection{Task 2: wall interaction reasoning}
Task 2 is the most polarized: some models sustain wall-collision reasoning across all horizons while others never develop a stable model of boundary reflection.

\begin{itemize}
    \item \textbf{Stable boundary reasoning:} \textit{GPT-5.5} leads with 90.34\% mean accuracy, \textit{Qwen3.6-Plus} follows at 87.39\%; both remain strong across 1--5 s rather than only at short windows. \textit{GPT-5.4-Pro} also remains in the top tier, showing that strong Task 2 performance is shared by several of the best overall models.
    
    \item \textbf{Long-horizon specialization:} \textit{Claude-Opus-4.6} attains the best 5 s score (89.54\%) despite not ranking near the top overall. \textit{GPT-5.4} and \textit{Claude-Opus-4.7} remain weak on Task 2, confirming that terminal-state prediction alone does not translate into high overall rank.
\end{itemize}

\subsection{Task 3: precise coordinate estimation}
Task 3 requires exact $[x, y]$ coordinates for all balls, measuring whether a model can maintain numerically grounded predictions rather than only classify events.
\begin{itemize}
    \item \textbf{Top performance:} \textit{GPT-5.4-Pro} and \textit{GPT-5.5} reach 87.93\% and 86.93\% at 1 s and remain above 60\% at 5 s. Precise final-state prediction is substantially stronger in these models than in earlier baselines.
    
    \item \textbf{Different axis from event reasoning:} \textit{GPT-5.4} and \textit{Claude-Opus-4.7} are strong on Task 3 despite weak event reasoning, whereas \textit{Qwen3.6-Plus} shows the reverse. The benchmark thus distinguishes predicting a plausible final state from identifying the events that causally produce it.
\end{itemize}

\subsection{Temporal dynamics and biases}
Longer horizons do not affect all models equally. Extending from 1 s to 5 s introduces longer interaction chains, greater friction effects, and more elastic reflections. Balanced models (\textit{GPT-5.5}, \textit{GPT-5.4-Pro}) degrade gradually; uneven models mainly expose existing weaknesses rather than creating new ones.

Failures are non-uniform: \textit{GPT-5.4} and \textit{Claude-Opus-4.7} miss event-level questions even when final-state estimates remain strong, while \textit{Qwen3.6-Plus} loses accuracy on terminal-state prediction. This is consistent with ``stasis bias'': under growing uncertainty, some models default to conservative event judgments even when their terminal-state predictions remain plausible. The benchmark becomes more diagnostic as the horizon increases, not merely harder.

\subsection{Overall physical reasoning performance}
\begin{figure*}[t]

\includegraphics[width=\textwidth]{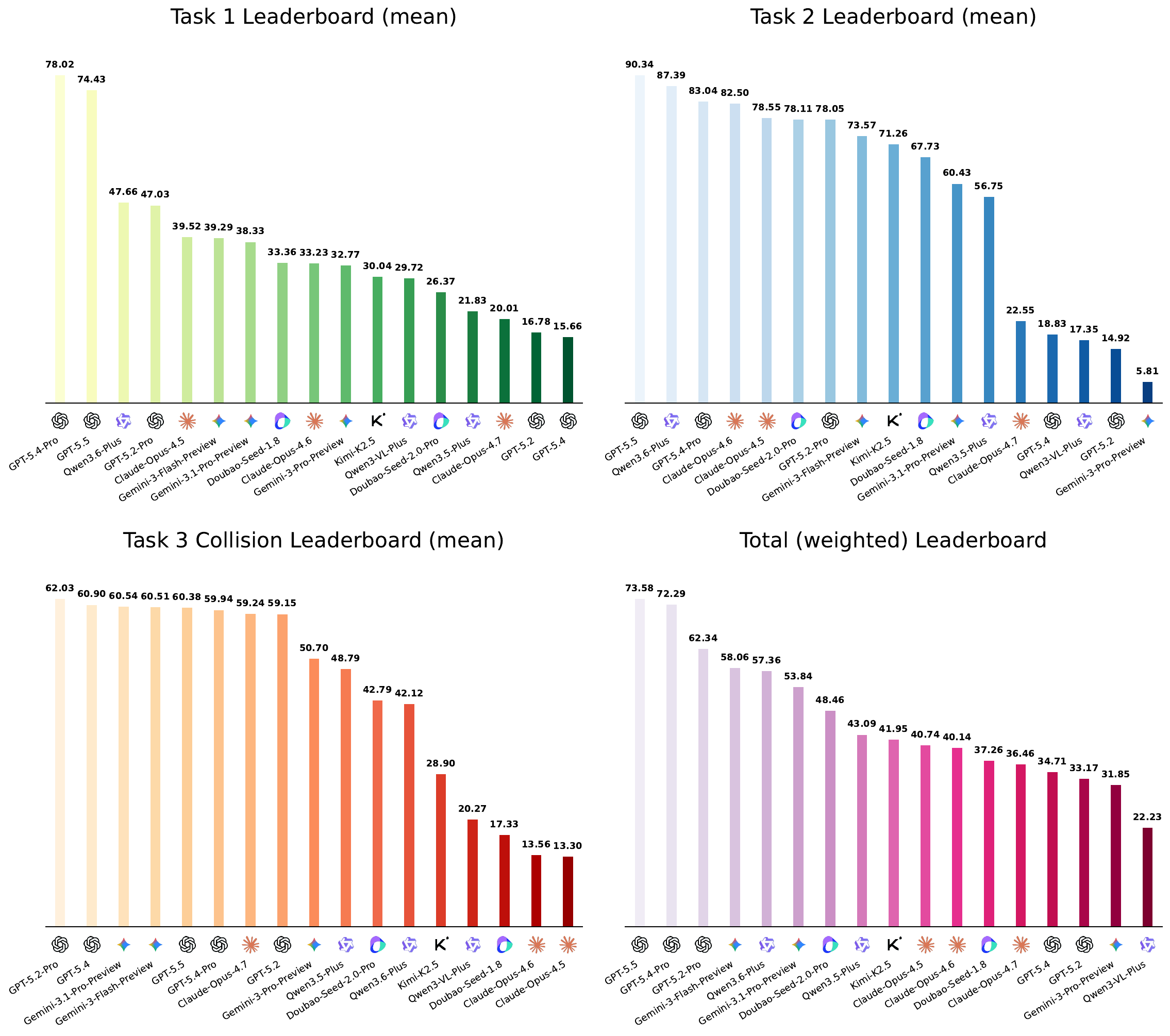}

\caption{\textbf{Aggregate performance leaderboards across the diagnostic tasks.} The bar charts show mean accuracy averaged over all temporal horizons (1 s to 5 s). (Top-left) \textbf{Task 1 leaderboard}: mean accuracy for discrete collision prediction. (Top-right) \textbf{Task 2 leaderboard}: mean accuracy for wall-interaction reasoning. (Bottom-left) \textbf{Task 3 leaderboard}: mean accuracy for precise 2D coordinate estimation. (Bottom-right) \textbf{Total (weighted) leaderboard}: overall ranking based on a weighted combination of the three tasks.}\label{fig:totalscore}
\end{figure*}

The total score (Figure~\ref{fig:totalscore}, bottom-right) combines the three tasks:
\begin{equation}
    S_{total} = 0.3 \cdot Acc_{T1} + 0.3 \cdot Acc_{T2} + 0.4 \cdot Acc_{T3}
\end{equation}
Task 3 receives the highest weight (0.4) because it requires precise coordinate prediction and is the most demanding.

The leaderboard rewards models that avoid a sharp trade-off between event reasoning and terminal-state prediction. \textit{GPT-5.5} ranks first (73.58\%), with \textit{GPT-5.4-Pro} close behind (72.29\%); their profiles differ slightly (\textit{GPT-5.5} strongest on Task 2, \textit{GPT-5.4-Pro} more balanced). \textit{GPT-5.2-Pro} is the strongest earlier baseline (62.34\%). Below this group, \textit{Gemini-3-Flash-Preview} and \textit{Qwen3.6-Plus} remain competitive---the former balanced, the latter carried by event reasoning. \textit{GPT-5.4} and \textit{Claude-Opus-4.7} confirm that strong terminal-state prediction alone is insufficient when event reasoning is weak.

\section{Discussion}

\subsection{The GPT family: model progression and test-time compute}
The GPT family contains both cross-generation changes (\textit{GPT-5.2} $\to$ \textit{GPT-5.4} $\to$ \textit{GPT-5.5}) and within-generation reasoning-depth variants (base vs.\ Pro). The ``Pro'' variants are deeper-reasoning configurations rather than separate base models.

Additional reasoning depth helps most on event inference. This is visible in the \textit{GPT-5.2}/\textit{GPT-5.2-Pro} comparison, where Task 1 and Task 2 improve far more than terminal-state prediction. It becomes clearer in the \textit{GPT-5.4}/\textit{GPT-5.4-Pro} pair: under the lighter setting, \textit{GPT-5.4} is ``terminal-state strong, event-level weak''; under deeper reasoning, \textit{GPT-5.4-Pro} becomes one of the most balanced models. This suggests that event-level weakness in base configurations can be mitigated by more test-time reasoning.

Cross-generation change is not monotonic. Moving from \textit{GPT-5.2} to \textit{GPT-5.4} does not produce uniform gain; instead the model becomes more polarized, with stronger terminal-state prediction but persistently weak event reasoning. \textit{GPT-5.5} then improves on the high-reasoning regime represented by \textit{GPT-5.4-Pro}, achieving the best overall score and Task 2 performance, but its advantage is incremental rather than a restructuring of capability.

A cost trade-off: \textit{GPT-5.4-Pro} and \textit{GPT-5.5} consume roughly $8\times$ more tokens than other models including \textit{GPT-5.4}, with most of the extra budget going to reasoning tokens. Because prompt and client settings are held fixed, this gap reflects model-side default reasoning behavior rather than something introduced by our evaluation setup. We cannot inspect hidden reasoning traces directly, so this conclusion is based on usage metadata. This does not diminish the strongest GPT results, but accuracy and inference efficiency are not aligned in this comparison.

\subsection{The Gemini family}
\textit{Gemini-3-Flash-Preview} (58.06\%) and \textit{Gemini-3.1-Pro-Preview} (53.84\%) offer balanced profiles, remaining competitive on terminal-state prediction but clearly behind the top GPT models on event reasoning and aggregate score. The Gemini family shows useful consistency across tasks, yet consistency alone is no longer enough to secure top ranks once stronger event-reasoning models are included.

\subsection{The Qwen series}
\textit{Qwen3.6-Plus} (57.36\%) far exceeds \textit{Qwen3.5-Plus} (43.09\%) and \textit{Qwen3-VL-Plus} (22.23\%). Its profile is distinctive:
\begin{itemize}
    \item \textbf{Strong event reasoning:} Task 2 at 87.39\% (stable between 84.92\% and 91.29\% across horizons), competitive on Task 1 at 47.66\%.
    \item \textbf{Weaker terminal-state prediction:} Task 3 drops from 57.00\% at 1 s to 44.71\% at 5 s, suggesting better local event judgments than long-horizon physical rollout.
\end{itemize}
Although it does not match the best GPT models overall, it clearly belongs to the second tier and represents a different route to strong performance: robust event reasoning without equally strong final-state prediction.

\subsection{The Claude family: an internal split}
The Claude family's progression is non-monotonic. \textit{Claude-Opus-4.5} and \textit{Claude-Opus-4.6} are strongest on wall-interaction reasoning (\textit{4.6}: 82.50\% Task 2, best 5 s score at 89.54\%). They retain meaningful boundary-reflection strength even when overall scores are mid-range.

\textit{Claude-Opus-4.7} inverts this: strong on Task 3 (77.93\% at 1 s, 62.27\% at 5 s) but weak on event reasoning (20.01\% Task 1, 22.55\% Task 2). Like \textit{GPT-5.4}, it is ``terminal-state strong, event-level weak.'' This model also has a small number of missing samples. The Claude line does not show steady improvement; later variants shift the balance between reflection reasoning and terminal-state prediction rather than uniformly improving.

\subsection{Mid-range models: Kimi and Doubao}
\textit{Doubao-Seed-2.0-Pro} (48.46\%) gains over \textit{Doubao-Seed-1.8} (37.26\%) mainly through better Task 2 (67.73\% $\to$ 78.11\%) and Task 3 (21.07\% $\to$ 49.03\%), while Task 1 drops (33.36\% $\to$ 26.37\%). The confusion statistics are consistent with a more conservative prediction style: fewer false positives but more missed true collisions. Its stronger overall result comes from wall-interaction reasoning and terminal-state prediction, not from collision-event detection.

\textit{Kimi-K2.5} (41.95\%) occupies a different middle position with a more even profile: 30.04\% Task 1, 71.26\% Task 2, 33.53\% Task 3. It does not match the Task 2/3 performance of \textit{Doubao-Seed-2.0-Pro} but avoids the severe weaknesses of \textit{Doubao-Seed-1.8}, which is limited by poor \texttt{final.json} completeness beyond 3 s. Within this group, \textit{Doubao-Seed-2.0-Pro} is the most competitive and \textit{Kimi-K2.5} the most balanced.

\subsection{Implications for physical world modeling}
Physical reasoning is not a single ability. Different model families show different trade-offs between terminal-state prediction and event reasoning. Strong performance requires both reliable spatial grounding and a mechanism for reasoning about causal interactions over time. Models strong in only one dimension---\textit{GPT-5.4} and \textit{Claude-Opus-4.7} on spatial grounding, \textit{Qwen3.6-Plus} on event reasoning---do not achieve top overall scores even when individual task performance is high.

\subsection{Modeling vs.\ calculation}
Some models approximate where objects end up (\textit{GPT-5.4}: 80.86\% Task 3 at 1 s) without identifying the collision events that produce that outcome (25.76\% Task 1). \textit{Qwen3.6-Plus} shows the reverse: strong on Task 2 across all horizons but weaker on Task 3.

Task 3 can sometimes be approached as terminal-state extrapolation from the observed scene. Task 1 requires identifying state transitions---collisions that redirect trajectories. The strongest models connect event-level reasoning with long-horizon prediction rather than excelling at only one. This gap suggests some models are better at producing spatially plausible outputs than at representing the causal structure of the physical process. Improving this connection is important for physical reasoning in multimodal systems.

\section{Conclusion}
\label{sec:conclusion}

We introduced \texttt{BilliardPhys-Bench}, a benchmark for evaluating physical reasoning in MLLMs through procedurally generated billiard scenarios. The results reveal a consistent gap between recognizing a scene and reasoning about how it will evolve.

\textbf{GPT-5.5} and \textbf{GPT-5.4-Pro} form the leading tier, with \textbf{GPT-5.5} achieving the strongest overall score and \textbf{GPT-5.4-Pro} the most balanced profile. \textbf{Qwen3.6-Plus} forms a strong second tier on event reasoning. \textbf{GPT-5.4} and \textbf{Claude-Opus-4.7} show that terminal-state prediction alone is insufficient without reliable event detection.

Performance declines as simulation time increases from 1 s to 5 s: small errors in early trajectory estimation accumulate, and longer interaction chains become harder to predict. Current MLLMs still lack a stable model of physical dynamics over time. \texttt{BilliardPhys-Bench} provides a direct way to evaluate these limitations and measure progress toward models with stronger physical inductive biases.

\section*{Limitations}
\label{sec:limitations}

Our benchmark uses a constant-friction model with perfectly elastic collisions in two dimensions. Real billiards involve spin, non-uniform friction, imperfect elasticity, and 3D dynamics such as jump shots. The simplified setting isolates core reasoning abilities but does not capture all of real-world billiard physics.

All scenarios are billiard-table scenes. Generalization to other physical domains (fluid dynamics, deformable objects, articulated bodies) cannot be assumed from performance on this benchmark alone.

The benchmark evaluates prediction from a single initial-state image. Models receive no video sequences or multiple frames, so the benchmark tests forward simulation from a static input rather than dynamic state estimation.

We did not collect human performance data. A human baseline would help contextualize model results and separate tasks that are inherently difficult from those that expose model-specific weaknesses.

Most evaluated models are accessed via commercial APIs whose architectures, training data, and parameter counts are not publicly disclosed. This limits reproducibility and detailed failure analysis.

All scenes contain exactly seven balls (IDs 0--6) on a standard table. The benchmark does not vary the number of objects, table geometry, or obstacle configurations.


\bibliography{references}

\end{document}